\pdfoutput=1

\documentclass[11pt]{article}

\usepackage[final]{acl}

\usepackage{times}
\usepackage{latexsym}
\usepackage{braket}
\usepackage{amssymb}
\usepackage{amsthm}
\usepackage{amsmath}

\usepackage[T1]{fontenc}

\usepackage[utf8]{inputenc}

\usepackage{microtype}

\usepackage{inconsolata}

\usepackage{graphicx}
\usepackage{xcolor}
\usepackage{microtype}
\usepackage{hyperref}
\usepackage{url}
\usepackage{booktabs}
\usepackage{amsmath}
\usepackage{pifont}
\usepackage{enumitem}
\usepackage{multirow}
\usepackage{array}

\DeclareMathOperator*{\argmin}{arg\,min}

\newcommand{\Table}[1]{Table~\ref{#1}}

\newcommand{\Appendix}[1]{Appendix~\ref{#1}}

\newcommand{\Figure}[1]{Fig.~\ref{#1}}

\newcommand{\Sectref}[1]{Section~\ref{#1}}

\newcommand{\Sref}[1]{\S\ref{#1}}

\newcolumntype{L}[1]{>{\raggedright\let\newline\\\arraybackslash\hspace{0pt}}m{#1}}
\newcolumntype{C}[1]{>{\centering\let\newline\\\arraybackslash\hspace{0pt}}m{#1}}
\newcolumntype{R}[1]{>{\raggedleft\let\newline\\\arraybackslash\hspace{0pt}}m{#1}}

\title{Locating Information Gaps and Narrative Inconsistencies Across Languages: A Case Study of LGBT People Portrayals on Wikipedia}

\author{
Farhan Samir$^{1,4}$\thanks{Work done while visiting the University of Washington.}~~~Chan Young Park$^{2}$~~~Anjalie Field$^{3}$\AND Vered Shwartz$^{1,4}$~~~Yulia Tsvetkov$^{2}$\\
$^1$ University of British Columbia\qquad
$^2$ University of Washington\\ 
$^3$ Johns Hopkins University\qquad
$^4$ Vector Institute for AI\\
\texttt{fsamir@cs.ubc.ca}}

\newcommand{\xfactret}{\textsc{X-FactAlign}}
\newcommand{\xfactmatch}{\textsc{X-FactMatch}}
\newcommand{\infogap}{\textsc{InfoGap}}
\newcommand{\lgbtbiocorpus}{\textsc{LGBTBioCorpus}}

\begin{document}
\maketitle
\begin{abstract}

To explain social phenomena and identify systematic biases, much research in computational social science focuses on comparative text analyses. 
These studies often rely on coarse corpus-level statistics or local word-level analyses, mainly in English.
We introduce the \infogap{} method---an efficient and reliable approach to \emph{locating information gaps and inconsistencies in articles  
at the fact level, across languages.} We evaluate \infogap{} by analyzing LGBT people's portrayals, across 2.7K biography pages on English, Russian, and French Wikipedias. We find large discrepancies in factual coverage across the languages. Moreover, our analysis reveals that 
biographical facts carrying negative connotations are more likely to be highlighted in Russian Wikipedia. Crucially, \infogap{} both facilitates large scale analyses, and pinpoints local document- and fact-level information gaps, laying a new foundation for targeted and nuanced comparative language analysis at scale.\footnote{\url{https://github.com/smfsamir/infogap}}

\end{abstract}

\section{Introduction}
\label{sec:intro}
\begin{figure}[t]    \centering\includegraphics[width=\columnwidth]{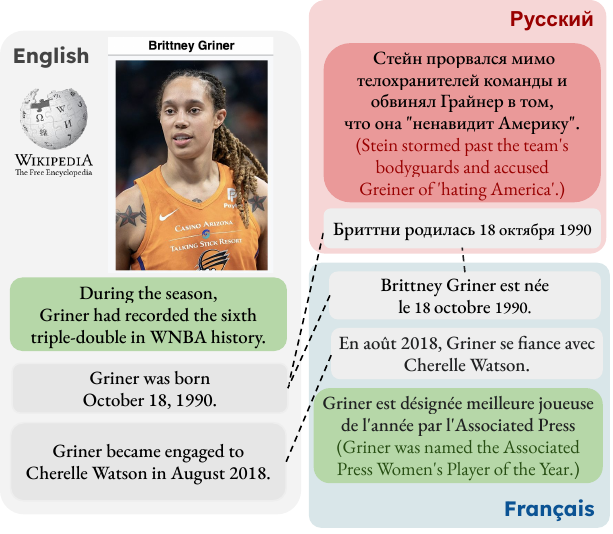}
    \caption{{We propose a method, \infogap{}, to locate fact (mis)alignments in  Wikipedia biographies in different language versions. \infogap{} identifies facts that are common to a pair of articles (``Griner was born on October 18, 1990''), and facts unique to one language version (``Griner had recorded the sixth triple-double''; \texttt{En} only) enabling further analysis of information gaps, editors' selective preferences within articles, and analyses at scale across languages, cultures, and demographics. }}
    \label{fig:infogap-label}
    \vspace{-15pt}
\end{figure}


Wikipedia has several hundred language editions, a sizeable number of which have more than 100K articles. 
Despite its ``neutral point of view'' policy, abundant evidence of content discrepancies across language editions  has been well-documented on the platform \cite[e.g.,][]{hecht2010tower,callahan2011cultural,eom2015interactions,wagner2015s,park2021multilingual}.
There are numerous motivations for identifying and studying these variations, e.g., identifying content variations and gaps can aid editors in removing 
social and cultural biases
\cite{field2022controlled}. 
Alternatively, from a social science perspective, comparative analyses of prominent topics across Wikipedia language editions provides a window into studying cross-cultural differences at scale \citep{callahan2011cultural}.


Existing methods for examining cross-language differences and gaps across Wikipedias rely on aggregate statistics, such as the number of languages an article is available in \citep{wagner2015s}, summary metrics of positive and negative connotations \citep{park2021multilingual}, text complexity measures \citep{kim2016understanding,field2022controlled}, or differences in hyperlink graph structures \citep{hecht2010tower,laufer2015mining}. While these metrics are useful for understanding broad trends, they do not facilitate nuanced comparative analysis, failing to inform readers and editors \emph{how} the content they engage with varies across language versions.
At the same time, manual fine-grained comparative analyses \citep[e.g., ][]{callahan2011cultural} do not scale and 
run the risk of incorporating researchers' biases.

In this work, we propose \infogap{}, a highly reliable method for identifying overlaps and gaps across different language articles on the same topic. Our method is composed of two steps: an \emph{alignment} step aimed at aligning facts across different language versions, followed by a \emph{validation} step aimed at determining fact equivalence. 
\infogap{} allows us to automatically identify exact content differences, as illustrated in \Figure{fig:infogap-label}, enabling both aggregate and fine-grained comparative analyses (\Sref{sec:infogap}).

After verifying the accuracy of \infogap{} against our manual annotations, we demonstrate its usefulness through a comparative analysis on thousands of multilingual Wikipedia biographies from the \lgbtbiocorpus{} \citep{park2021multilingual}. We find that the coverage of public figures differs substantially across languages (\Sref{sec:lgbt_bio}). For example, when comparing Russian and English biographies, we find that on average $34\%$ of the content in Russian biographies is not present in their English counterparts. 

Critically, as suggested in manual analyses by \citet{park2021multilingual}, our 
automatic analyses at scale identify that many of the bios carry a significantly different implied sentiment towards the figure, depending on the language version that is accessed.\footnote{We focus on the LGBT subset of LGBTQIA+ people on Wikipedia, due to data scarcity for other groups.} Aggregating these sentiment imbalances across $2.7$K biographies, we contribute the insight that Russian LGBT biographies share disproportionately more negative sentiment facts with English biographies than positive ones. Overall, \infogap{} enables the pinpointing of fine-grained factual and framing distinctions between narratives across languages, aggregates these insights across thousands of articles, and  offers tools to identify the specific documents that most clearly highlight these nuances.

\section{\infogap{}: Identifying  Information Asymmetry in Wikipedia Articles} 
\label{sec:infogap}
\begin{figure}
    \centering\includegraphics[width=0.85\columnwidth]{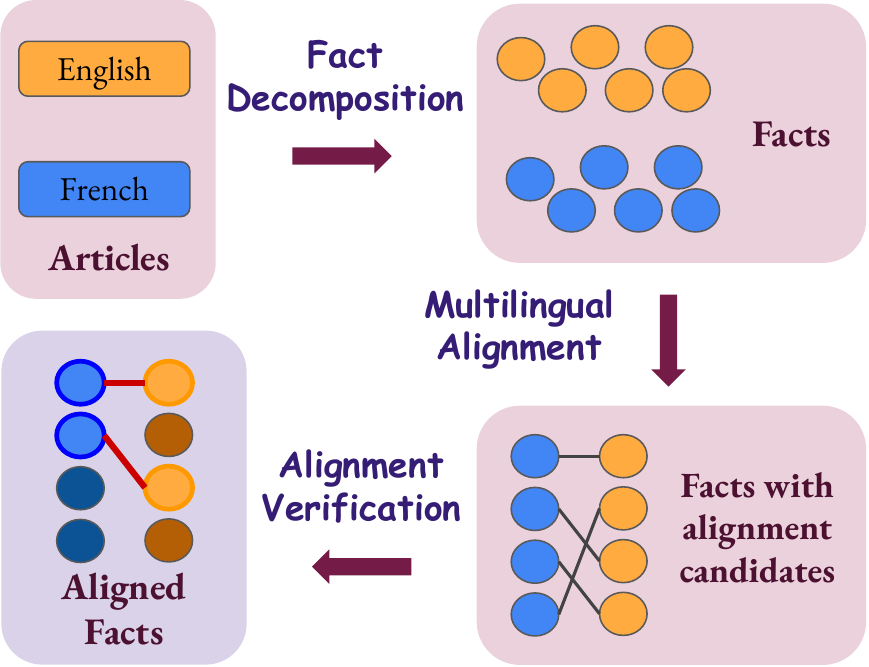}
    \caption{Schematic of the \infogap{} procedure. We describe the Fact Decomposition and Multilingual Alignment steps in \Sref{sec:x-fact-retrieve},  and  the Alignment Verification step in \Sref{sec:x-fact-eq}.}
    \label{fig:schematic}
\end{figure}

Consider a pair of articles on a topic written in different languages. We call one article $E$ and the other in the pair $F$. Moreover, we represent $E$ by a series of facts $e_{1},\dots,e_{n}$ and $F$ similarly as $f_{1},\dots,f_{m}$. Our method determines for a given fact $e_{i}\in E$ whether it appears in $F$ ($F \Vdash e_i$) or not ($F\not\Vdash e_i$). The pipeline is directional, so we can compute both $F\Vdash e_i$ and $E\Vdash f_i$. Without loss of generality, we will describe the procedure for obtaining the labels $F\vdash e_i$, for all $e_i$. We refer to this as the $E\to F$ direction.   
 
 \Figure{fig:schematic} presents an overview of \infogap{}. We primarily focus on two steps. First, following \newcite{min-etal-2023-factscore}, we narrow the search space of equivalent facts by aligning a fact in $E$ to facts in $F$ that may convey the same information (Sec~\ref{sec:x-fact-retrieve}). This allows us to efficiently assess the equivalence between aligned facts (Sec~\ref{sec:x-fact-eq}). We determine the reliability of \infogap{} in Section~\ref{sec:method:reliability}. 

\subsection{\xfactret: Cross-Lingual~Fact~Alignment} 
\label{sec:x-fact-retrieve}

\paragraph{Fact Decomposition.} As a first step, we need to represent an article (e.g. $E$) as a series of facts ($e_{1},\dots,e_{n}$). Sentences are suboptimal for this purpose, as they can be overly complex. Instead, following \citet{kamoi-etal-2023-wice}, we use GPT-4 \citep{achiam2023gpt} for fact decomposition. Differently from \citet{kamoi-etal-2023-wice}, who decompose sentences, we decompose entire paragraphs, to provide more context to the model and allow it to resolve co-references. See Appendix~\ref{sec:appendix:fact_decomposition} for the prompt. 

\paragraph{Fact Representation.} In order to determine whether $e_{i}$ is also conveyed in $F$, we embed each fact in $E$ and in $F$ using multilingual LaBSE embeddings \citep{Feng2020LanguageagnosticBS}. The straightforward way to align facts is by computing the cosine similarity between $e_{i}$ and each fact $f_j \in F$, aligning $e_{i}$ to the most similar fact: $\argmin_{j}d(\mathbf{e_i}, \mathbf{f_j})$. We find this approach can further be improved by considering the context of the surrounding facts. In the following paragraphs, we describe two improvements we made in \xfactret. First, we restrict the pool of paragraphs in $F$ from which $f_j$ can be retrieved. Second, we apply an adjustment to the computation of $d(\cdot, \mathbf{f_i})$, accounting for the hubness of $\mathbf{f_j}$ \citep{lazaridou2015hubness}.    

\paragraph{Paragraph Alignment.} We can partition the facts in $E$ into their paragraphs: $P_{E}^1, ..., P_{E}^N$. Similarly for $F$: $P_{F}^1, ..., P_{F}^M$. We represent each paragraph by the set of its facts' embeddings.  We then construct a bipartite graph between paragraphs in $E$ and paragraphs in $F$, adding a directed edge from each paragraph in $E$,  $P_{E}^i$, to a paragraph in $F$,  $P_{F}^{j}$ such that $j = \text{\texttt{MaxSim} }_j\: d(P_{E}^i, P_F^{j})$ \citep{khattab2020colbert}. We do the same in the other direction, going from $F$ to $E$. Removing the direction from the edges, we obtain an adjacency matrix $A$ between the paragraphs so that each paragraph in $E$ is connected to at least one paragraph in $F$. For a given fact $e_i$, we can now limit the pool of alignment candidates in $F$ to $f_j$s where the paragraphs of $e_i$ and $f_j$ are adjacent in the graph. 

\paragraph{Correcting for Hubness.}  Given that we are comparing facts from articles on the same topic, 
directly computing $d(\mathbf{e_i}, \mathbf{f_j})$ can lead to aligning unrelated facts that discuss the same common named entities. In particular, some facts $f_j$ are similar to many other facts $e_i$, causing a ``hubness problem'' \cite{lazaridou2015hubness, conneau2017word}.  To mitigate this, we follow \citet{artetxe-schwenk-2019-margin} and normalize $d(\cdot, \mathbf{f_j})$ so that it is a function of the semantic density of $\mathbf{f_j}$.  The density-normalized distance $D(\mathbf{e_i}, \mathbf{f_i})=d(\mathbf{e_i}, \mathbf{f_j})-\text{hubness}(\mathbf{f_j})$. We compute the hubness of $f_j$ by computing the average nearest neighbor distance ($k_{NN}=5$) between $f_j$ and 50 other facts drawn from paragraphs that are \emph{not} in the adjacency list of the paragraph containing $e_i$.   
Overall, this process enables us to retrieve $k=2$ facts from $F$ that may convey the same information as $e_i$. 

\subsection{\xfactmatch: Cross-Lingual~Fact~Matching}
\label{sec:x-fact-eq}

With $e_i$ and its aligned facts $f_j$, we can now answer the question whether a given fact $e_{i}\in E$ appears in $F$ ($F \Vdash e_i$) or not ($F\not\Vdash e_i$). We assume that if $F \Vdash e_i$, there exist facts in $F$ that entail $e_i$. In particular, we can expect these facts to be aligned with $e_i$. We thus relax the problem of judging whether $F\Vdash e_{i}$ to whether any of the facts $f_j$ retrieved by \xfactret{} entail $e_i$, i.e. whether $\operatorname{any}(\Set{f_j \Vdash e_{i} | j \in [k]})$.\footnote{We use $[k]$ for $\{1,\dots,k\}$; see \citet[][p. 11]{harvey2022first}.}

We use entailment as a shorthand for ``conveying the same information as'' despite a minor deviation from the definition of entailment in linguistics as a strict logical entailment \citep{heim1998semantics}, and in NLP as ``a human [reading the premise] would typically think that the hypothesis is likely
true'' \citep{dagan2005pascal,bowman-etal-2015-large}. Our definition is a bit more relaxed and we also consider partial entailment \cite{levy-etal-2013-recognizing}, i.e., when the most important information in $e_i$ is conveyed by $F$, allowing the omission of peripheral information. To that end, we don't use existing NLI models. Furthermore, research on cross-language entailment detection is limited \citep{negri-etal-2012-semeval,Rodriguez2023XPARADECT}, and to our knowledge there are no publicly available models that can determine the entailment between a premise and a hypothesis in different languages. 

Inspired by \newcite{min-etal-2023-factscore} and \newcite{shafayat2024multi} who used GPT-4 to assess the truthfulness of a model-generated fact against a trusted knowledge base, we prompt GPT-4 to compare an English fact to its aligned facts in $F$. 
Concretely, we prompt the model with the hypothesis fact $e_i$ and the two immediately preceding facts for context ($e_{i-1}$ and $e_{i-2}$), along with all of the premise facts $f_j$ and their contexts ($f_{j-1}$ and $f_{j-2}$). We instruct the model to determine whether $e_{i}$ can be inferred from any of the $f_j$ ($j\in [k]$).  Appendix~\ref{sec:appendix:entailment_prompt} presents the prompt that we use for all language pairs. The model's prediction serves as the final label for whether $F\Vdash e_i$.


\subsection{Assessing the Reliability of \infogap{}}
\label{sec:method:reliability}


\begin{table}[t]
    \centering
    \begin{tabular}{L{2cm}L{1.7cm}R{1.3cm}}
        \toprule
        \textbf{Language Pair} & \textbf{\#Labeled } & \textbf{\#Annotated}  \\
        \midrule
        En $\to$ Fr. & 2,213 & \multirow{2}{*}{80}  \\
        Fr $\to$ En. & 2,165 &  \\
        \midrule
       En $\to$ Ru & 2,832 & \multirow{2}{*}{80}  \\
       Ru $\to$ En & 2,435 & \\
       \bottomrule
    \end{tabular}
    \caption{\textbf{Number of facts labeled using \infogap{}} for each language pair and direction, and number of manually annotated facts. }
    \label{tab:infogap-number-of-facts}
    \vspace{-10pt}
\end{table}

To assess the reliability of \infogap{}, we evaluate its final results with human annotations. We apply \infogap{} to Wikipedia biographies in English and French of 10 people, and in English and Russian for 12 people, comprising nearly 10K facts altogether. We draw on biographies from the \textsc{LGBTBioCorpus} \citep{park2021multilingual}, a corpus we analyze in \Sref{sec:lgbt_bio} at a larger scale. See  \Table{tab:infogap-number-of-facts} for a breakdown of the number of facts and Appendix~\ref{appendix:biographies} for the biographies.

We annotated a subset of the facts in each language pair and direction. 
Given a hypothesis $e_i$, the retrieved candidate facts $f_j$ from \xfactret{}, and their contexts, we ask the annotator to choose between three options: (1) the hypothesis fact $e_i$ can be inferred from one of the retrieved $f_j$; (2) the hypothesis fact can be inferred from the article $F$, but not from the $f_j$ (indicating that \xfactret{} failed to retrieve the correct fact); (3) $e_i$ cannot be inferred from $F$. We also provide relaxed versions of options (1) and (2), where $e_{i}$ can be partially inferred from one of the retrieved $f_j$ or partially inferred from $F$.
Concretely, our annotation task closely resembles the \xfactmatch{} step, with two key differences. First, we provide the annotators with English translations of non-English facts and their contexts, using the NLLB model \citep{costa2022no}. Second, if a hypothesis fact $e_i$ cannot be inferred from the $k$ facts retrieved by \xfactret{}, we ask the annotator to read the full Wikipedia article $F$ to determine whether $e_i$ can be inferred from it. 

One author annotated $80$ facts in each language pair and for both directions within each language pair. Another author annotated 40 of those 80 for each language pair. We obtained substantial inter-annotator agreements, with Cohen's $\kappa=0.71$ for \texttt{En}/\texttt{Fr} and $\kappa=0.78$ for \texttt{En}/\texttt{Ru}. We thus conclude that the task is relatively unambiguous. 

\begin{table}
    \centering
    \small
    \begin{tabular}{L{1.5cm}lll}
        \toprule
        \textbf{Language pair} & \textbf{\infogap{}} & \textbf{NLI} & \textbf{Random}\\
        \midrule
        \texttt{En $\to$ Fr} & $\mathbf{0.81}$ & $0.28$ & $0.62$ \\
          \texttt{Fr $\to$ En} & $\mathbf{0.90}$ & $0.27$ & $0.61$ \\
        \midrule
        \texttt{En $\to$ Ru} & $\mathbf{0.78}$ & $0.52$ & $0.43$ \\ 
        \texttt{Ru $\to$ En} & $\mathbf{0.88}$ & $0.50$ & $0.35$ \\
        \bottomrule
    \end{tabular}
    \caption{Performance of \infogap{} with respect to the manual annotations ($n=80$ for each language pair), in terms of $F_1$ score.} 
    \label{tab:infogap-f1s}
\end{table}

In order to determine the reliability of \infogap{}, we compute the predictions against the annotated $80$ facts for each language pair. \Table{tab:infogap-f1s} presents the $F_1$ scores that range from 0.78 to 0.9, indicating that the \infogap{} pipeline is highly reliable in identifying whether a fact in $E$ is present in $F$ (and vice-versa). Substituting \xfactmatch{} with a RoBERTa NLI  baseline \citep{liu2019roberta} performs significantly worse.\footnote{The baseline is available on HuggingFace as \texttt{cross-encoder/nli-roberta-base}.} The RoBERTa NLI model rarely predicts an entailment label on our dataset, resulting in its poor performance. With the exception of \texttt{En $\to$ Ru}, the NLI baseline is outperformed by a classifier that randomly predicts whether the target fact is entailed. \infogap{} significantly outperforms both \citep[$p<0.05$, with a bootstrap percentile test;][]{efron1994introduction}.


\section{Using \infogap{} to Analyze Asymmetries in LGBT Wikipedia Bios}
\label{sec:lgbt_bio}
\newcommand{\entofr}{\texttt{En $\to$ Fr}}
\newcommand{\entoru}{\texttt{En $\to$ Ru}}
\newcommand{\rutoen}{\texttt{Ru $\to$ En}}
\newcommand{\frtoen}{\texttt{Fr $\to$ En}}
\newcommand{\en}{\texttt{En}}
\newcommand{\fr}{\texttt{Fr}} 
\newcommand{\ru}{\texttt{Ru}}

Having validated the effectiveness of \infogap{}, we move onto applying it to answer questions about information gaps in Wikipedia. We focus on identifying content differences between language versions' articles on LGBT public figures. Prior work by \citet{park2021multilingual} identified that English articles on average portrayed these figures with more positive sentiment, as well as greater power and agency \citep{sap2017connotation}, relative to articles in Russian and Spanish.

To gain further insight into cross-linguistic variation towards LGBT people portrayals, we draw on the \lgbtbiocorpus{} corpus \citep{park2021multilingual}. The corpus comprises $1,350$ biographies of LGBT people, each paired with biographies of non-LGBT people matched on most social attributes except sexual orientation using the matching method introduced in \citet{field2022controlled}. Given that \infogap{} enables us to directly compare the content between different language versions of a biography, we contend that our analysis can provide a more direct characterization of differences in LGBT bios. Specifically, we look at \en{}, \fr{}, and \ru{} Wikipedias.
We consider the following research questions:
\begin{itemize}[nosep,leftmargin=*,align=left,itemindent=9pt]
    \item[\textbf{RQ$_1$}:] To what extent does factual knowledge differ across language versions of the same bio (Sec~\ref{sec:lgbt-rq1})?
    \item[\textbf{RQ$_2$}:] Does a person's affiliation with the LGBT community have an effect on the information gap in their bios (Sec~\ref{sec:lgbt-rq2})? 
    \item[\textbf{RQ$_3$}:] Can we use \infogap{} to identify sections to remediate (Sec~\ref{sec:lgbt-rq3})? 
\end{itemize}

These questions are intentionally ordered from high-level (language-level) to low-level (individual- and fact-level) to demonstrate that \infogap{} enables both high-level quantitative analyses and efficient low-level descriptive analyses.
\noindent We start by providing the implementation details in Section~\ref{sec:lgbt-implementation} before answering each of the research questions. 

\subsection{Implementation Details}
\label{sec:lgbt-implementation}

The \lgbtbiocorpus{} is significantly larger than the small set of $22$ biographies from \Sectref{sec:infogap}, leading to high cost and runtime when applying \infogap{}. Parsing Alan Turing's biography with \infogap{} alone, for example, can require more than 100K GPT-4 tokens.\footnote{We used \texttt{gpt-4-1106-preview}. At the time of writing, this API cost $\$30.00\text{ USD} / 1$M tokens.} We thus use the GPT-4 predictions to finetune smaller models that are more efficient.    
Specifically, we use \texttt{flan-t5-large} \citep{chung2024scaling} for both directions of the \en{}/\fr{} pair and \texttt{mt5-large} \citep{xue2020mt5} for both directions of the \en{}/\ru{} pair.
We find that the T5 variants perform well at modeling the annotations from \Sref{sec:infogap}, obtaining macro-averaged F1 scores of $0.90$ (\entoru{}, \rutoen{}) and $0.87$ (\entofr{}, \frtoen{}). We provide fine-tuning hyperparameters and validation set performances in Appendix~\ref{sec:appendix:hyperparams}. 

\subsection{RQ$_1$: Information Gaps in Bios}
\label{sec:lgbt-rq1}
\begin{figure}
    \centering
    \includegraphics[scale=0.50]{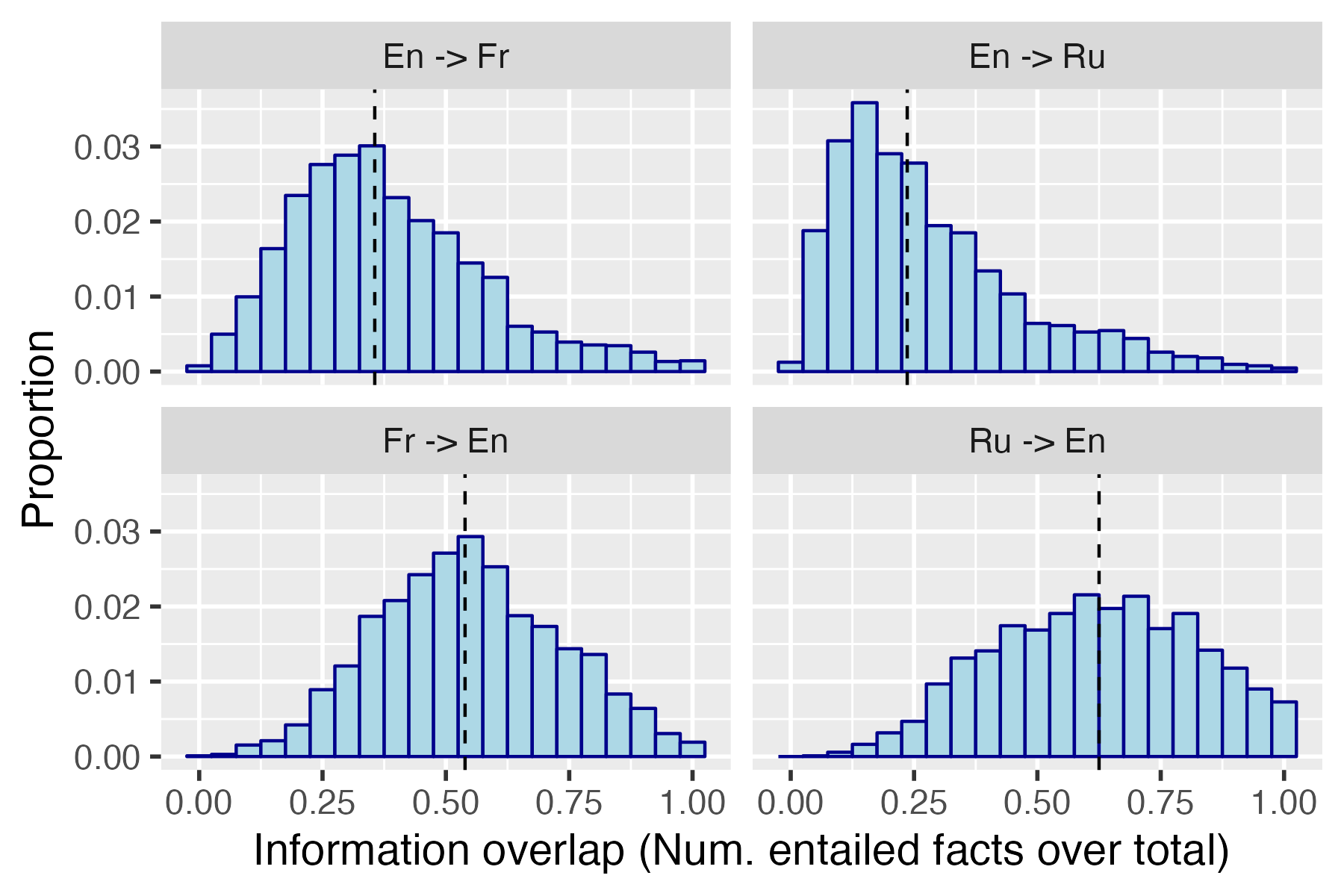}
    \caption{\textbf{Distribution of information overlaps for LGBTBioCorpus}. Top: Distribution over the percentage of facts in \en{} biographies also found in their \fr{} and \ru{} counterparts. Bottom: Distribution over the percentage of facts in  \fr{} and \ru{} biographies also found in their English counterparts. $N=2,700$ biographies. 
    In general, \en{} biographies contain more facts that are exclusive to \en{}.}
    \label{fig:overlap-distribution}
    \vspace{-10pt}
\end{figure}
Previous smaller-scale manual qualitative analyses showed that people portrayals differ systematically across language versions \citep{callahan2011cultural}. However, this was challenging to quantify as it would be unreasonably laborious to manually count the number of overlapping facts between language versions of an article.   
Equipped with \infogap{}, we can for the first-time quantify variance in information overlap between language versions of Wikipedia biographies at scale.
Specifically, we consider the \infogap{} predictions for the entire corpus (LGBT and non-LGBT bios). 

\Figure{fig:overlap-distribution} visualizes the distribution of the number of facts that can be found in both language versions of the same bio. In the top-left subfigure, we show a histogram of the amount of information in the \en{} article that can also be found in the \fr{} article (\entofr{}). The median of the distribution is $0.35$, indicating that for half of the biographies, only 35\%  of the information in the \en{} article can be found in the \fr{} article.  By comparison, the median of the \texttt{Fr $\to$ En} distribution is 0.55, much higher than the median of the \entofr{} distribution, indicating that \en{} biographies contain more unique information than their \fr{} counterparts. 

 Considering \en{}/\ru{}, we find that \en{} articles contain significantly more unique information than \ru{} counterparts, with the median \entoru{} overlap being $0.23$. Much of the information in the \ru{} articles meanwhile can be found in the \en{} articles, with a median overlap of $0.66$ for \rutoen{}. 

We also note that the \infogap{} ratios reflect the well known ``local heros'' effect, where biographies of individuals whose nationality matches the language of the article tend to have greater coverage, length, and visibility \citep{callahan2011cultural,field2022controlled,hecht2010tower,Oeberst2024HowAC}. When the nationality of the person is Russian (66 people), the  median \texttt{En $\to$ Ru} overlap increases to $0.29$ ($+5\%$) while the \texttt{Ru $\to$ En} overlap decreases to 0.44 ($-22\%$). Similarly for French (148 people), the median \texttt{En $\to$ Fr} overlap increases to 0.52 ($+17\%$), while the \texttt{Fr $\to$ En} overlap decreases to 0.29 ($-26\%$). 
Overall, this result indicates that there are large scale disparities in information overlap ratios across language versions, building on \citeauthor{callahan2011cultural}'s (\citeyear{callahan2011cultural}) early analysis.

\subsection{RQ$_2$: Effect of LGBT Affiliation on Information Gaps}
\label{sec:lgbt-rq2}
\begin{table}[t]
    \centering
    \small
    \begin{tabular}{clr}
        \textbf{Language pair} & \textbf{Factor} & \textbf{Coefficient} \\
        \toprule
        \texttt{En $\to$ Ru} & \texttt{conn\_pos}** & -0.16 \\  
        & \texttt{conn\_neg}** & -0.18\\
        & \texttt{is\_lgbt}** & 0.10\\
        & \texttt{conn\_pos:is\_lgbt} & 0.04\\
        & \texttt{conn\_neg:is\_lgbt} & 0.03\\
        \midrule
        \texttt{En $\to$ Fr} & \texttt{conn\_pos}** & -0.07 \\  
        & \texttt{conn\_neg} & 0.01\\
        & \texttt{is\_lgbt}** & -0.05\\
        & \texttt{conn\_pos:is\_lgbt} & 0.01\\
        & \texttt{conn\_neg:is\_lgbt}** & 0.06\\
        \midrule
        \rutoen & \texttt{conn\_pos}** & -0.14 \\  
        & \texttt{conn\_neg}** & -0.51\\
        & \texttt{is\_lgbt}** & 0.26\\
        & \texttt{conn\_pos:is\_lgbt} & 0.04\\
        & \texttt{conn\_neg:is\_lgbt}** & 0.25\\
        \midrule
        \frtoen & \texttt{conn\_pos}** & -0.07 \\  
        & \texttt{conn\_neg}** & -0.14\\
        & \texttt{is\_lgbt} & 0.00\\
        & \texttt{conn\_pos:is\_lgbt} & 0.03\\
        & \texttt{conn\_neg:is\_lgbt}** & 0.09\\
        \bottomrule
    \end{tabular}
    \caption{\textbf{Mean of posterior distribution of regression coefficients}. ** indicates that 95\% posterior credible interval for the coefficient does \textit{not} contain zero. }
    \label{tab:regression-coefficients}
    \vspace{-10pt}
\end{table}


Given the large scale differences in content between language versions, we turn to the question of whether LGBT people biographies exhibit different patterns of information overlap compared to non-LGBT people. For example, do Russian biographies tend to include or exclude certain types of information depending on whether the biography is about an LGBT person? To investigate this question, we fit a binomial regression model to determine which factors contribute to the inclusion of \en{} facts in the corresponding \fr{} or \ru{} bios, and vice versa.



\paragraph{Features.} Table~\ref{tab:regression-coefficients} displays the features we use, along with their coeffcient estimates.\footnote{We also fit models with the covariates of \texttt{gender}, \texttt{nationality}, and \texttt{ethnicity}. Including these covariates did not change the estimates for the features in \Table{tab:regression-coefficients}, so we omit them for clarity.} Naturally, we include a binary feature \texttt{is\_lgbt} indicating whether the bio is of an LGBT person. Crucially, we also need to consider the connotation of facts in the English article. \citet{park2021multilingual} found that English LGBT bios were portrayed with greater sentiment, power, and agency than Russian bios. However, this prior work cannot shed light on whether the difference in sentiment is due to Russian bios including negative sentiment facts that are not in the English bios, excluding positive sentiment facts from the English bios, or both. We directly address this question using \infogap{}. 

To determine the connotation of a fact $e_i$, we have to consider the context in its original sentence, which requires mapping between a fact and a sentence. To map facts to their original sentences (e.g., ``\textit{Cook is on the board of directors of Nike}'' $\to$ ``\textit{Cook is also on the boards of directors of Nike, Inc. and the National Football Foundation}''), we use forced alignment; see \Appendix{appendix:connotation}. 

We obtain connotation predictions at the sentence level by prompting a language model to determine whether a given sentence (in the context of the two prior sentences) portrays the subject of the biography in a positive, negative, or neutral light. Similar to the distillation of the \infogap{} process to a smaller model (\Sref{sec:lgbt-implementation}),  we first obtain connotation labels using GPT-4 for a smaller set of bios, and use those labels to finetune a smaller model for scaling to the full \textsc{LGBTBioCorpus} (see Appendix~\ref{appendix:connotation} for details, including human annotation of the connotation labels). We use the sentence-level connotation label as the label for its constituent facts. \Table{tab:connotation-distribution} presents this label distribution. 


\begin{table}[t]
    \centering
    \small
    \begin{tabular}{llll}
         \textbf{Language} & \textbf{Positive} & \textbf{Neutral} & \textbf{Negative} \\
         \toprule
         English & $0.434$ & $0.488$ & $0.077$\\ 
         French & $0.442$ & $0.455$ & $0.102$\\ 
         Russian & $0.327$ & $0.658$ & $0.014$ \\
         \bottomrule
    \end{tabular}
    \caption{\textbf{Distribution of implied sentiment} about biography subjects for \en{}, \fr{}, and \ru{} articles.}
    \label{tab:connotation-distribution}
    \vspace{-5pt}
\end{table}
\paragraph{Regression Model.} 
Without loss of generality, consider modeling the amount of information in the \en{} bios that is also present in the \fr{} bios, i.e., the \entofr{} direction.
To perform our binomial regression, we first partition each bio into three sets -- positive, negative, and neutral facts. Each partition represents one datapoint for fitting the regression model, so each bio contributes three datapoints. Within each of these three partitions, some facts will also be present in $F$, while others will be exclusive to $E$. We model this using a bayesian binomial regression model \citep{mcelreath2018statistical}:
{
\setlength{\abovedisplayskip}{5pt}
\setlength{\belowdisplayskip}{5pt}
\setlength{\abovedisplayshortskip}{0pt}
\setlength{\belowdisplayshortskip}{0pt}
\begin{equation*}
\small
    \text{\texttt{overlap}} \mid 
N_p \sim 1 + \text{\texttt{conn}} +  \text{\texttt{is\_lgbt}} + \text{\texttt{is\_lgbt}:\texttt{conn}}
\end{equation*}
}

\noindent where {\texttt{conn} gets the value of either \texttt{conn\_pos}, \texttt{conn\_neg}, or \texttt{conn\_neutral}, depending on the input partition, $N_p$ is the number of facts in the current partition of $E$, and \texttt{overlap} is the number of facts that are also in $F$ (at most $N_p$). \texttt{is\_lgbt}:\texttt{conn} is an interaction between the two categorical variables. See \Appendix{appendix:bayesian-regression} for model-fitting details.  

\paragraph{Connotation is a predictive factor.} Listed in \Table{tab:regression-coefficients}, our results indicate that connotation is a predictive factor 
in nearly all language pairs and directions considered, except \texttt{conn\_neg} in \entofr{}. Further, the polarity of the \texttt{conn\_pos} and \texttt{conn\_neg} factors is always negative, suggesting that polarized facts tend to be included in lower rates than neutral facts, which are more agreeable across language versions. To ground the effect size of the coefficients, we can simulate predictions from the regression model. For example, a value of -0.07 for \texttt{conn\_pos} in the \entofr{} model indicates that 34.4\% of the positive  facts in \en{} are included in the \fr{} bios, compared to 36.6\% of the neutral facts. 

\paragraph{Negative connotation facts are disproportionately included in Russian LGBT bios.} Considering Russian biographies, we draw from the large coefficient value of the \texttt{is\_lgbt} feature that facts from the English article are more likely to be referenced when the article is about an LGBT public figure. Moreover, from the \texttt{is\_lgbt:conn\_neg} interaction, we find that negative facts are more likely to be referenced than positive ones.
To quantify the size of this effect, we simulate posterior predictions from the binomial regression model. We find an average 50.87\% of negative Russian facts are shared with the English biographies when they describe an LGBT public figure, whereas only 38.53\% of negative facts are shared with English bios when they are non-LGBT.

\subsection{RQ$_3$: Identifying Sections to Remediate}
\label{sec:lgbt-rq3}

Our analysis in Sec~\ref{sec:lgbt-rq2} revealed that facts carrying a more polarizing (non-neutral) connotations are less likely to be shared across language versions. This suggests that many biographies may carry a significantly different overall connotation, depending on the language version in which they are read. Unlike the manual analysis performed in prior works \citep[e.g.,][among others]{park2021multilingual,callahan2011cultural} to identify such language-version imbalanced content, \infogap{} can automatically locate imbalanced content. \citet{park2021multilingual} in particular focused on bios where the subject was portrayed with a more negative implied sentiment. Here, we focus on a different aspect of sentiment differences: the \textit{omission} of content with positive implied sentiment from one language version.

Specifically, we follow these steps to identify imbalanced content: First, we identify bios from the \lgbtbiocorpus{} where a high rate of positive facts are excluded from one language version compared to another.\footnote{It is also effective at identifying individual facts present in one language version but absent in another (see \Sref{sec:infogap}), but for this analysis we consider collections of multiple facts.} Next, we introduce a method to identify positive life events that are missing in that language version. We provide a formal argument demonstrating that our \infogap{} based method for identifying missing events is highly accurate. Finally, we conclude with examples of findings.


\paragraph{Step 1. Identifying biographies with imbalanced implied sentiment.} 
Consider a pair of articles $E$ and $F$ written in different languages, and suppose we wanted to find bios where $F$ omitted positive content at a high rate.  We conduct a hypothesis test to determine whether the number of positive facts included in both languages  is significantly lower than expected based on the overlap rate of neutral facts. Concretely, we perform a bayesian hypothesis test based on the \texttt{BetaBinomial} distribution; we provide complete details in \Appendix{appendix:imbalanced-bio-test}.

Our test identifies $274$ imbalanced LGBT biographies when considering \entoru{} and $236$ when considering \entofr{}. We can follow the same procedure for finding English biographies that comparatively lack positive information, when compared to their French and Russian counterparts. We find 105 and 199 biographies in the \rutoen{} and \frtoen{} direction, respectively.

\paragraph{Step 2. Identifying events that are unique to a language version.} Having identified biographies that could benefit from remediation, we next focus on finding the positive-connotation carrying content that is missing from one language version. By comparison to \citet{park2021multilingual}, who could only analyze $10$ biographies for identifying imbalanced content, we can leverage \infogap{} to identify imbalanced content at scale within the subset of biographies we identified. We focus on finding positive connotation \textit{events} -- longer collections of facts that are thematically related -- rather than individual isolated facts since the omission of a whole event is more egregious. Practically speaking, we search for paragraphs $\mathcal{V} = e_1, \dots, e_{N_V}$ where all facts in the paragraph are missing from $F$:
{
\setlength{\abovedisplayskip}{5pt}
\setlength{\belowdisplayskip}{5pt}
\setlength{\abovedisplayshortskip}{0pt}
\setlength{\belowdisplayshortskip}{0pt}
\begin{equation}
 M = \{ \mathcal{V} \in E | \operatorname{all}(\Set{F \not \Vdash e_{i} | i \in [N_V]})\}  
 \label{eq:missing_events}
\end{equation}
}
We then select a subset of $M$: paragraphs containing at least one positive connotation fact.



\paragraph{\infogap{} is highly effective at identifying missing events.} Consider an event $\mathcal{V} = e_1, \dots, e_{N_V}$ that is described in article $E$. Suppose that \infogap{} predicted that $\mathcal{V}$ is not covered by $F$, that is that $F$ does not entail any of the events in $\mathcal{V}$: $F\not \vDash e_i, i\in[N_V]$. For \infogap{} to be wrong, i.e., $\mathcal{V}$ is actually present in $F$, there needs to exist a subset of facts $e_{i(1)}, \dots, e_{i(k)} \in \mathcal{V}$, for $k=p\cdot N_V$ ($0<p<1$), that are entailed by $F$: $F \vDash e_{i(j)}, j\in[k]$. We can bound the probability of this error. 

\newtheorem{prop}{Proposition}
\newtheorem{theorem}{Theorem}
\newtheorem{remark}{Remark}

\begin{prop}[Error Bound of Event Identification through InfoGap]
The probability of \infogap{} making $k$ errors is $\leq \exp(-2(1-\epsilon)^2k)$, where $\epsilon$ is the error rate of the classifier when it predicts $F\not \vDash e_i$.  
\end{prop}
\newcommand{\present}{\textcolor{teal}{\checkmark}}

\newcommand{\xmark}{\textcolor{red}{\ding{55}}}

\begin{table*}[t]
    \centering
    \small
    \begin{tabular}{llp{4.5in}}
         \textbf{Pair} & \textbf{Person} & \textbf{Events} \\
         \toprule
          \en{} \xmark, \ru{} \present & Tim Cook & In 2022, following Russia's invasion of Ukraine, \textbf{Tim Cook} called on the company's employees to donate to help Ukraine. \textbf{Apple's CEO} announced the decision to suspend sales of equipment in Russia and also said that the company would triple the amount of donations made by employees to support Ukraine, and this would be retroactive to February 25, 2022.\\
         \midrule
         \en{} \xmark, \fr{} \present & Chelsea Manning & ``Ron Paul, a leader of the libertarian movement within the Republican Party, endorsed \textbf{Manning} on April 12, 2013, stating that \textbf{Manning} had done more for peace than Obama—referring to Obama's 2009 Nobel Peace Prize win: ``While President Obama was initiating and expanding unconstitutional wars abroad, \textbf{Manning}, whose actions caused exactly zero deaths, was shining a light on the truth behind those wars. Which of the two has done more for peace is clear.''\\ 
\midrule
         \en{} \present, \fr{} \xmark & Caster Semenya & In 2010, the British magazine \textit{New Statesman} included \textbf{Semenya} in its annual list of ``50 People That Matter'' for unintentionally instigating ``an international and often ill-tempered debate on gender politics, feminism, and race, becoming an inspiration to gender campaigners around the world'' \\
         \midrule
         \en{} \present, \ru{} \xmark & Ada Colau & During her period as mayor of Barcelona, \textbf{Colau} has maintained a political stance against activities that are susceptible of contributing to greenhouse gas emissions and air pollution. She has repeatedly opposed the expansion of El Prat airport and the use of private cars in the city, and has pushed regional authorities to restrict the number of cruise ships arrivals in Barcelona. In 2020 she declared a ``climate emergency'', advocating limiting the consumption of meat at schools and forbidding councillors from using the Barcelona-Madrid air shuttle.\\
    \bottomrule
    \end{tabular}
    \caption{Examples of events from biographies that contain a large number of positive facts that are only contained in one language version of the article relative to another. We provide translations (Google Translate) for the first two rows, rather than the original French and Russian content.}
    \label{tab:imbalanced-examples}
\end{table*}

\begin{proof}
Given that the error rate of the classifier is $\epsilon$, the expected number of errors for $k$ predictions is $\epsilon \cdot k$. However, the classifier made $k$ mistakes, so we have made $\epsilon \cdot k + (1-\epsilon)\cdot k$ errors, an additive factor of $t = (1-\epsilon)\cdot k$ more mistakes than expected. By Hoeffding's inequality (\Appendix{appendix:hoeffding}), where we supply our expected value $\mu=\epsilon \cdot k$ and the deviation from the expected value of $t=(1-\epsilon)\cdot k$, we obtain an upper bound of:
{
\setlength{\abovedisplayskip}{2pt}
\setlength{\belowdisplayskip}{2pt}
\setlength{\abovedisplayshortskip}{0pt}
\setlength{\belowdisplayshortskip}{0pt}
\begin{equation*}
    \leq \exp\left(-2(1-\epsilon)^{2}k^2/k\right) = \exp(-2(1-\epsilon)^2 k).
\end{equation*}
}
\end{proof}
The significance of this claim is that it is rare for the \infogap{} classifier to make a large number ($k$) of mistakes when the error rate is $\epsilon$ (where $\epsilon << 1$). Moreover, the probability  of mistakes decreases very quickly in the accuracy of the classifier and the number of facts in $\mathcal{V}$ that were predicted to not be entailed by $F$. As we showed empirically in \Sectref{sec:infogap}, the \infogap{} classifier is reliable (low $\epsilon$) and thus it has a strong capacity to find events that are only described in one language version.\footnote{One shortcoming of this argument is if $F$ discusses a completely different aspect of the event $\mathcal{V}$ than $E$. We conjecture that this is unlikely since both articles should at least contain the central propositions about the event.} 

\paragraph{Findings.} In \Table{tab:imbalanced-examples}, we demonstrate positive events that are unique to one language version when compared to another. We find that Chelsea Manning's \fr{} page describes praise for her whistleblowing during the Afghanistan war. The \fr{} page also discusses her whistleblowing on the Abu Ghraib prison conditions \citep{hersh2004abu}. Conspicuously, both events are omitted from the \en{} page, despite the \en{} page being otherwise longer. American perception of this instance of whistleblowing skewed negative \citep{pew2010wikileaks}, which may have played a role in the disparities between the \en{} and \fr{} pages.

We also find Tim Cook's \ru{} page -- but not his \en{} page -- makes note of his fundraising initiative to defend Ukraine in the current Russo-Ukranian war. It is unsurprising that it appears in the \ru{} page, as it directly pertains to Russia. However, the omission of this fact from the \en{} page is remarkable, since it had received some media attention from American outlets \citep{verge2022email}. One reason for this omission may be that there is a partisan divide on US involvement in the war \citep{pew2024ukraine}. This fact may not have been included in \en{} to maintain a veneer of neutrality.

It is important to note however that Wikipedia's Neutral Point of View policy advocates for a balanced representation of views \citep{matei2011wikipedia}, rather than outright filtering or censorship. Our findings raise questions about the degree to which a cross-linguistically consistent ``Neutral Point of View''  is realizable. \infogap{} enables studying these cross-linguistic differences in portrayals of public figures at scale.



\section{Related Work} 
\label{sec:related_work}
\paragraph{Automated comparison of multilingual Wikipedia articles.} We contribute to a large body of work on understanding differences between language versions of Wikipedia. \citet{hecht2010tower} also compare Wikipedia language versions and consider their information gaps, and later develop a web tool to bridge these multilingual gaps \citep{bao2012omnipedia}. However, their evaluation is at a higher level of abstraction: they look at whether or not two language versions have on a topic. By comparison, we compare content differences between two language versions on the same topic.  \citet{duh2013managing} considered a pipeline similar to \infogap{} for the task of keeping multilingual Wikipedia documents consistent. However, their pipeline used embedding similarity; in early experiments, we found that using embedding similarity for identifying potential entailments performed very poorly (relative to \xfactmatch{}). \citet{massa2012manypedia} created a web tool that permits visual comparison of Wikipedia articles in two different languages. 
\citet{Rodriguez2023XPARADECT} also perform comparative analyses across language versions in Wikipedia. However, they consider more fine-grained content differences between pairs of the most closely related paragraphs between different language versions' article on a topic. Their method was not designed for computing the overall article level overlaps and differences of the form we demonstrate in \Figure{fig:overlap-distribution}.

\paragraph{Case studies on cultural differences in multilingual Wikipedia.} We highlight two studies that were not mentioned elsewhere in this work. \citet{hickman2021understanding} analyze how a boundary dispute over Kashmir between India and Pakistan is represented in English, Hindi, and Urdu Wikipedia, analyzing how the Neutral Point of View principle is upheld. They find there is a sizeable number of cross-language editors between Urdu and English, as well as Hindi and English, but not Urdu and Hindi, attributing this to the popularity of English Wikipedia. \citet{kharazian2024governance} studied how the Croatian language version of Wikipedia was usurped by a small group of editors who aimed to promote far-right bias and disinformation about various Croatian political figures, groups, and events. This bias was apparent when comparing the Croatian articles to Serbian and English ones.

\section{Conclusion}
\label{sec:conclusion}
We presented \infogap{}, a reliable method for efficient comparative analysis between two narratives on the same topic written in different languages. We deployed the method to discover  differences in LGBT people's portrayals, locating shared facts, as well as information gaps and inconsistencies across 2.7K English, Russian, and French Wikipedia biography pages. 
\infogap{} can be directly applied beyond analyzing differences in multilingual Wikipedia biographies. Analyzing variation in topic coverage is at the heart of much research in the social sciences, from understanding media manipulation strategies \citep{field-etal-2018-framing}, to analyzing differences in argumentation from different stances in a contentious debate \citep{luo-etal-2020-detecting}, to analyzing quotation patterns in partisan media \citep{niculae2015quotus}. 
Overall, our research lays the foundation 
for enabling targeted, nuanced textual comparative analyses at scale.

\section{Limitations}
\label{sec:limitations}
\paragraph{Applicability to specialized domains.} Our method relies on the language understanding abilities of the underlying language model (GPT-4 in our case). While we were able to achieve high accuracy on the \lgbtbiocorpus{}, it is not guaranteed that similarly high-accuracy can be achieved if we were to apply \infogap{} to more specialized domains, where domain expertise may be required to assess the equivalence of two facts in different languages, such as comparing Wikipedia articles concerning scientific topics. 

\paragraph{Connotation is subjective.} In Section~\ref{sec:lgbt_bio} we investigated the effect of connotation on the inclusion of facts in different language versions. We acknowledge that connotation is fairly subjective, and may depend on a reader's stance towards the topic and their cultural background. To ensure a high degree of replicability of our results, we have released our all of the finetuned models we applied in \Sref{sec:lgbt_bio}, including the connotation models.


\paragraph{Ablations of \infogap{} components.} We did not perform ablations of the components of the \xfactret{} step in the \infogap{} pipeline (\Sref{sec:infogap}). Our aim was to demonstrate that high-quality automatic cross-lingual comparative analysis is not only possible (\Sref{sec:method:reliability}) but provides considerable benefits in downstream analyses (\Sref{sec:lgbt_bio}). We will perform thorough ablations with a larger number of annotated samples in future work. 


\section{Ethical Considerations}
\label{sec:ethical_considerations}
\paragraph{Data.} The dataset used in this study, \lgbtbiocorpus{},  is publicly available. 



\paragraph{Models.} We used language models to make classification predictions, limiting their ability to generate offensive content. We used a closed-source model, GPT-4, which entails high costs, and may not be suitable for applying our method to different datasets, especially those containing private information. The distilled version of \infogap{}, which uses open-source models, addresses both concerns.


\section{Acknowledgements} 
We thank Shreya Prakash for advice on our regression analyses and hypothesis testing. We thank Miikka Silfverberg and Jai Aggarwal for helpful feedback on the manuscript. FS is supported by an NSERC PGS-D scholarship. VS is supported by the Vector Institute for AI, the CIFAR AI Chair program, and NSERC. 
We gratefully acknowledge support from the National Science Foundation under CAREER Grant No.~IIS2142739, and NSF grants No.~IIS2125201 and IIS2203097.

\bibliography{custom,anthology}

\begin{thebibliography}{51}
\providecommand{\natexlab}[1]{#1}

\bibitem[{Achiam et~al.(2023)Achiam, Adler, Agarwal, Ahmad, Akkaya, Aleman, Almeida, Altenschmidt, Altman, Anadkat et~al.}]{achiam2023gpt}
Josh Achiam, Steven Adler, Sandhini Agarwal, Lama Ahmad, Ilge Akkaya, Florencia~Leoni Aleman, Diogo Almeida, Janko Altenschmidt, Sam Altman, Shyamal Anadkat, et~al. 2023.
\newblock Gpt-4 technical report.
\newblock \emph{arXiv preprint arXiv:2303.08774}.

\bibitem[{Artetxe and Schwenk(2019)}]{artetxe-schwenk-2019-margin}
Mikel Artetxe and Holger Schwenk. 2019.
\newblock \href {https://doi.org/10.18653/v1/P19-1309} {Margin-based parallel corpus mining with multilingual sentence embeddings}.
\newblock In \emph{Proceedings of the 57th Annual Meeting of the Association for Computational Linguistics}, pages 3197--3203, Florence, Italy. Association for Computational Linguistics.

\bibitem[{Bao et~al.(2012)Bao, Hecht, Carton, Quaderi, Horn, and Gergle}]{bao2012omnipedia}
Patti Bao, Brent Hecht, Samuel Carton, Mahmood Quaderi, Michael Horn, and Darren Gergle. 2012.
\newblock Omnipedia: bridging the wikipedia language gap.
\newblock In \emph{Proceedings of the SIGCHI Conference on Human Factors in Computing Systems}, pages 1075--1084.

\bibitem[{Bowman et~al.(2015)Bowman, Angeli, Potts, and Manning}]{bowman-etal-2015-large}
Samuel~R. Bowman, Gabor Angeli, Christopher Potts, and Christopher~D. Manning. 2015.
\newblock \href {https://doi.org/10.18653/v1/D15-1075} {A large annotated corpus for learning natural language inference}.
\newblock In \emph{Proceedings of the 2015 Conference on Empirical Methods in Natural Language Processing}, pages 632--642, Lisbon, Portugal. Association for Computational Linguistics.

\bibitem[{B{\"u}rkner(2017)}]{burkner2017brms}
Paul-Christian B{\"u}rkner. 2017.
\newblock brms: An r package for bayesian multilevel models using stan.
\newblock \emph{Journal of statistical software}, 80:1--28.

\bibitem[{Callahan and Herring(2011)}]{callahan2011cultural}
Ewa~S Callahan and Susan~C Herring. 2011.
\newblock Cultural bias in wikipedia content on famous persons.
\newblock \emph{Journal of the American society for information science and technology}, 62(10):1899--1915.

\bibitem[{Chung et~al.(2024)Chung, Hou, Longpre, Zoph, Tay, Fedus, Li, Wang, Dehghani, Brahma et~al.}]{chung2024scaling}
Hyung~Won Chung, Le~Hou, Shayne Longpre, Barret Zoph, Yi~Tay, William Fedus, Yunxuan Li, Xuezhi Wang, Mostafa Dehghani, Siddhartha Brahma, et~al. 2024.
\newblock Scaling instruction-finetuned language models.
\newblock \emph{Journal of Machine Learning Research}, 25(70):1--53.

\bibitem[{Clark and Schiffer(2022)}]{verge2022email}
Mitchell Clark and Zoë Schiffer. 2022.
\newblock Read {T}im {C}ook’s email to employees on {U}kraine.
\newblock \emph{The Verge}.

\bibitem[{Conneau et~al.(2017)Conneau, Lample, Ranzato, Denoyer, and J{\'e}gou}]{conneau2017word}
Alexis Conneau, Guillaume Lample, Marc'Aurelio Ranzato, Ludovic Denoyer, and Herv{\'e} J{\'e}gou. 2017.
\newblock Word translation without parallel data.
\newblock \emph{arXiv preprint arXiv:1710.04087}.

\bibitem[{Costa-juss{\`a} et~al.(2022)Costa-juss{\`a}, Cross, {\c{C}}elebi, Elbayad, Heafield, Heffernan, Kalbassi, Lam, Licht, Maillard et~al.}]{costa2022no}
Marta~R Costa-juss{\`a}, James Cross, Onur {\c{C}}elebi, Maha Elbayad, Kenneth Heafield, Kevin Heffernan, Elahe Kalbassi, Janice Lam, Daniel Licht, Jean Maillard, et~al. 2022.
\newblock No language left behind: Scaling human-centered machine translation.
\newblock \emph{arXiv preprint arXiv:2207.04672}.

\bibitem[{Dagan et~al.(2005)Dagan, Glickman, and Magnini}]{dagan2005pascal}
Ido Dagan, Oren Glickman, and Bernardo Magnini. 2005.
\newblock The pascal recognising textual entailment challenge.
\newblock In \emph{Machine learning challenges workshop}, pages 177--190. Springer.

\bibitem[{Duh et~al.(2013)Duh, Yeung, Iwata, and Nagata}]{duh2013managing}
Kevin Duh, Ching-Man~Au Yeung, Tomoharu Iwata, and Masaaki Nagata. 2013.
\newblock Managing information disparity in multilingual document collections.
\newblock \emph{ACM Transactions on Speech and Language Processing (TSLP)}, 10(1):1--28.

\bibitem[{Efron and Tibshirani(1994)}]{efron1994introduction}
Bradley Efron and Robert~J Tibshirani. 1994.
\newblock \emph{An introduction to the bootstrap}.
\newblock Chapman and Hall/CRC.

\bibitem[{Eom et~al.(2015)Eom, Arag{\'o}n, Laniado, Kaltenbrunner, Vigna, and Shepelyansky}]{eom2015interactions}
Young-Ho Eom, Pablo Arag{\'o}n, David Laniado, Andreas Kaltenbrunner, Sebastiano Vigna, and Dima~L Shepelyansky. 2015.
\newblock \href {https://doi.org/10.1371/journal.pone.0114825} {Interactions of cultures and top people of wikipedia from ranking of 24 language editions}.
\newblock \emph{PloS one}, 10(3):1--27.

\bibitem[{Feng et~al.(2020)Feng, Yang, Cer, Arivazhagan, and Wang}]{Feng2020LanguageagnosticBS}
Fangxiaoyu Feng, Yinfei Yang, Daniel~Matthew Cer, N.~Arivazhagan, and Wei Wang. 2020.
\newblock \href {https://api.semanticscholar.org/CorpusID:220347683} {Language-agnostic bert sentence embedding}.
\newblock In \emph{Annual Meeting of the Association for Computational Linguistics}.

\bibitem[{Field et~al.(2018)Field, Kliger, Wintner, Pan, Jurafsky, and Tsvetkov}]{field-etal-2018-framing}
Anjalie Field, Doron Kliger, Shuly Wintner, Jennifer Pan, Dan Jurafsky, and Yulia Tsvetkov. 2018.
\newblock \href {https://doi.org/10.18653/v1/D18-1393} {Framing and agenda-setting in {R}ussian news: a computational analysis of intricate political strategies}.
\newblock In \emph{Proceedings of the 2018 Conference on Empirical Methods in Natural Language Processing}, pages 3570--3580, Brussels, Belgium. Association for Computational Linguistics.

\bibitem[{Field et~al.(2022)Field, Park, Lin, and Tsvetkov}]{field2022controlled}
Anjalie Field, Chan~Young Park, Kevin~Z Lin, and Yulia Tsvetkov. 2022.
\newblock Controlled analyses of social biases in wikipedia bios.
\newblock In \emph{Proceedings of the ACM Web Conference 2022}, pages 2624--2635.

\bibitem[{Harvey(2022)}]{harvey2022first}
Nick Harvey. 2022.
\newblock A first course in randomized algorithms.

\bibitem[{Hecht and Gergle(2010)}]{hecht2010tower}
Brent Hecht and Darren Gergle. 2010.
\newblock The tower of babel meets web 2.0: user-generated content and its applications in a multilingual context.
\newblock In \emph{Proceedings of the SIGCHI conference on human factors in computing systems}, pages 291--300.

\bibitem[{Heim and Kratzer(1998)}]{heim1998semantics}
Irene Heim and Angelika Kratzer. 1998.
\newblock \emph{Semantics in Generative Grammar}.
\newblock Blackwell.

\bibitem[{Hersh(2004)}]{hersh2004abu}
Seymour~M. Hersh. 2004.
\newblock \href {https://www.newyorker.com/magazine/2004/05/10/torture-at-abu-ghraib} {Torture at abu ghraib}.
\newblock \emph{New Yorker}.

\bibitem[{Hickman et~al.(2021)Hickman, Pasad, Sanghavi, Thebault-Spieker, and Lee}]{hickman2021understanding}
Molly~G Hickman, Viral Pasad, Harsh~Kamalesh Sanghavi, Jacob Thebault-Spieker, and Sang~Won Lee. 2021.
\newblock Understanding wikipedia practices through hindi, urdu, and english takes on an evolving regional conflict.
\newblock \emph{Proceedings of the ACM on Human-Computer Interaction}, 5(CSCW1):1--31.

\bibitem[{Hoffman et~al.(2014)Hoffman, Gelman et~al.}]{hoffman2014no}
Matthew~D Hoffman, Andrew Gelman, et~al. 2014.
\newblock The no-u-turn sampler: adaptively setting path lengths in hamiltonian monte carlo.
\newblock \emph{J. Mach. Learn. Res.}, 15(1):1593--1623.

\bibitem[{Kamoi et~al.(2023)Kamoi, Goyal, Diego~Rodriguez, and Durrett}]{kamoi-etal-2023-wice}
Ryo Kamoi, Tanya Goyal, Juan Diego~Rodriguez, and Greg Durrett. 2023.
\newblock \href {https://doi.org/10.18653/v1/2023.emnlp-main.470} {{W}i{CE}: Real-world entailment for claims in {W}ikipedia}.
\newblock In \emph{Proceedings of the 2023 Conference on Empirical Methods in Natural Language Processing}, pages 7561--7583, Singapore. Association for Computational Linguistics.

\bibitem[{Kharazian et~al.(2024)Kharazian, Starbird, and Hill}]{kharazian2024governance}
Zarine Kharazian, Kate Starbird, and Benjamin~Mako Hill. 2024.
\newblock Governance capture in a self-governing community: A qualitative comparison of the croatian, serbian, bosnian, and serbo-croatian wikipedias.
\newblock \emph{Proceedings of the ACM on Human-Computer Interaction}, 8(CSCW1):1--26.

\bibitem[{Khattab and Zaharia(2020)}]{khattab2020colbert}
Omar Khattab and Matei Zaharia. 2020.
\newblock Colbert: Efficient and effective passage search via contextualized late interaction over bert.
\newblock In \emph{Proceedings of the 43rd International ACM SIGIR conference on research and development in Information Retrieval}, pages 39--48.

\bibitem[{Kim et~al.(2016)Kim, Park, Hale, Kim, Byun, and Oh}]{kim2016understanding}
Suin Kim, Sungjoon Park, Scott~A Hale, Sooyoung Kim, Jeongmin Byun, and Alice~H Oh. 2016.
\newblock Understanding editing behaviors in multilingual wikipedia.
\newblock \emph{PloS one}, 11(5):e0155305.

\bibitem[{Laufer et~al.(2015)Laufer, Wagner, Fl{\"o}ck, and Strohmaier}]{laufer2015mining}
Paul Laufer, Claudia Wagner, Fabian Fl{\"o}ck, and Markus Strohmaier. 2015.
\newblock Mining cross-cultural relations from wikipedia: a study of 31 european food cultures.
\newblock In \emph{Proceedings of the ACM web science conference}, pages 1--10.

\bibitem[{Lazaridou et~al.(2015)Lazaridou, Dinu, and Baroni}]{lazaridou2015hubness}
Angeliki Lazaridou, Georgiana Dinu, and Marco Baroni. 2015.
\newblock Hubness and pollution: Delving into cross-space mapping for zero-shot learning.
\newblock In \emph{Proc. ACL}.

\bibitem[{Levy et~al.(2013)Levy, Zesch, Dagan, and Gurevych}]{levy-etal-2013-recognizing}
Omer Levy, Torsten Zesch, Ido Dagan, and Iryna Gurevych. 2013.
\newblock \href {https://aclanthology.org/P13-2080} {Recognizing partial textual entailment}.
\newblock In \emph{Proceedings of the 51st Annual Meeting of the Association for Computational Linguistics (Volume 2: Short Papers)}, pages 451--455, Sofia, Bulgaria. Association for Computational Linguistics.

\bibitem[{Liu(2019)}]{liu2019roberta}
Yinhan Liu. 2019.
\newblock Roberta: A robustly optimized bert pretraining approach.
\newblock \emph{arXiv preprint arXiv:1907.11692}.

\bibitem[{Luo et~al.(2020)Luo, Card, and Jurafsky}]{luo-etal-2020-detecting}
Yiwei Luo, Dallas Card, and Dan Jurafsky. 2020.
\newblock \href {https://doi.org/10.18653/v1/2020.findings-emnlp.296} {Detecting stance in media on global warming}.
\newblock In \emph{Findings of the Association for Computational Linguistics: EMNLP 2020}, pages 3296--3315, Online. Association for Computational Linguistics.

\bibitem[{MacKay(2003)}]{mackay2003information}
David~JC MacKay. 2003.
\newblock \emph{Information theory, inference and learning algorithms}.
\newblock Cambridge university press.

\bibitem[{Massa and Scrinzi(2012)}]{massa2012manypedia}
Paolo Massa and Federico Scrinzi. 2012.
\newblock Manypedia: Comparing language points of view of wikipedia communities.
\newblock In \emph{Proceedings of the Eighth Annual International Symposium on Wikis and Open Collaboration}, pages 1--9.

\bibitem[{Matei and Dobrescu(2011)}]{matei2011wikipedia}
Sorin~Adam Matei and Caius Dobrescu. 2011.
\newblock Wikipedia's “neutral point of view”: Settling conflict through ambiguity.
\newblock \emph{The Information Society}, 27(1):40--51.

\bibitem[{McElreath(2018)}]{mcelreath2018statistical}
Richard McElreath. 2018.
\newblock \emph{Statistical rethinking: A Bayesian course with examples in R and Stan}.
\newblock Chapman and Hall/CRC.

\bibitem[{Min et~al.(2023)Min, Krishna, Lyu, Lewis, Yih, Koh, Iyyer, Zettlemoyer, and Hajishirzi}]{min-etal-2023-factscore}
Sewon Min, Kalpesh Krishna, Xinxi Lyu, Mike Lewis, Wen-tau Yih, Pang Koh, Mohit Iyyer, Luke Zettlemoyer, and Hannaneh Hajishirzi. 2023.
\newblock \href {https://doi.org/10.18653/v1/2023.emnlp-main.741} {{FA}ct{S}core: Fine-grained atomic evaluation of factual precision in long form text generation}.
\newblock In \emph{Proceedings of the 2023 Conference on Empirical Methods in Natural Language Processing}, pages 12076--12100, Singapore. Association for Computational Linguistics.

\bibitem[{Mitzenmacher and Upfal(2017)}]{mitzenmacher2017probability}
Michael Mitzenmacher and Eli Upfal. 2017.
\newblock \emph{Probability and computing: Randomization and probabilistic techniques in algorithms and data analysis}.
\newblock Cambridge university press.

\bibitem[{Negri et~al.(2012)Negri, Marchetti, Mehdad, Bentivogli, and Giampiccolo}]{negri-etal-2012-semeval}
Matteo Negri, Alessandro Marchetti, Yashar Mehdad, Luisa Bentivogli, and Danilo Giampiccolo. 2012.
\newblock \href {https://aclanthology.org/S12-1053} {{S}emeval-2012 task 8: Cross-lingual textual entailment for content synchronization}.
\newblock In \emph{*{SEM} 2012: The First Joint Conference on Lexical and Computational Semantics {--} Volume 1: Proceedings of the main conference and the shared task, and Volume 2: Proceedings of the Sixth International Workshop on Semantic Evaluation ({S}em{E}val 2012)}, pages 399--407, Montr{\'e}al, Canada. Association for Computational Linguistics.

\bibitem[{Niculae et~al.(2015)Niculae, Suen, Zhang, Danescu-Niculescu-Mizil, and Leskovec}]{niculae2015quotus}
Vlad Niculae, Caroline Suen, Justine Zhang, Cristian Danescu-Niculescu-Mizil, and Jure Leskovec. 2015.
\newblock Quotus: The structure of political media coverage as revealed by quoting patterns.
\newblock In \emph{Proceedings of the 24th International Conference on World Wide Web}, pages 798--808.

\bibitem[{Oeberst and Ridderbecks(2024)}]{Oeberst2024HowAC}
Aileen Oeberst and Till Ridderbecks. 2024.
\newblock \href {https://api.semanticscholar.org/CorpusID:266842093} {How article category in wikipedia determines the heterogeneity of its editors}.
\newblock \emph{Scientific Reports}, 14.

\bibitem[{Park et~al.(2021)Park, Yan, Field, and Tsvetkov}]{park2021multilingual}
Chan~Young Park, Xinru Yan, Anjalie Field, and Yulia Tsvetkov. 2021.
\newblock Multilingual contextual affective analysis of lgbt people portrayals in wikipedia.
\newblock In \emph{Proceedings of the International AAAI Conference on Web and Social Media}, volume~15, pages 479--490.

\bibitem[{{Pew Research Center}(2010)}]{pew2010wikileaks}
{Pew Research Center}. 2010.
\newblock \href {https://www.pewresearch.org/politics/2010/12/08/most-say-wikileaks-release-harms-public-interest/} {Most say wikileaks release harms public interest}.
\newblock Technical report.

\bibitem[{{Pew Research Center}(2024)}]{pew2024ukraine}
{Pew Research Center}. 2024.
\newblock \href {https://www.pewresearch.org/global/2024/05/08/views-of-ukraine-and-u-s-involvement-with-the-russia-ukraine-war/} {Views of ukraine and u.s. involvement with the russia-ukraine war}.
\newblock Technical report.

\bibitem[{Rashkin et~al.(2015)Rashkin, Singh, and Choi}]{rashkin2015connotation}
Hannah Rashkin, Sameer Singh, and Yejin Choi. 2015.
\newblock Connotation frames: A data-driven investigation.
\newblock \emph{arXiv preprint arXiv:1506.02739}.

\bibitem[{Rodriguez et~al.(2023)Rodriguez, Erk, and Durrett}]{Rodriguez2023XPARADECT}
Juan~Diego Rodriguez, Katrin Erk, and Greg Durrett. 2023.
\newblock \href {https://api.semanticscholar.org/CorpusID:262043426} {X-parade: Cross-lingual textual entailment and information divergence across paragraphs}.
\newblock \emph{ArXiv}, abs/2309.08873.

\bibitem[{Sap et~al.(2017)Sap, Prasettio, Holtzman, Rashkin, and Choi}]{sap2017connotation}
Maarten Sap, Marcella~Cindy Prasettio, Ari Holtzman, Hannah Rashkin, and Yejin Choi. 2017.
\newblock Connotation frames of power and agency in modern films.
\newblock In \emph{Proceedings of the 2017 conference on empirical methods in natural language processing}, pages 2329--2334.

\bibitem[{Shafayat et~al.(2024)Shafayat, Kim, Oh, and Oh}]{shafayat2024multi}
Sheikh Shafayat, Eunsu Kim, Juhyun Oh, and Alice Oh. 2024.
\newblock Multi-fact: Assessing multilingual llms' multi-regional knowledge using factscore.
\newblock \emph{arXiv preprint arXiv:2402.18045}.

\bibitem[{Wagner et~al.(2015)Wagner, Garcia, Jadidi, and Strohmaier}]{wagner2015s}
Claudia Wagner, David Garcia, Mohsen Jadidi, and Markus Strohmaier. 2015.
\newblock It's a man's wikipedia? assessing gender inequality in an online encyclopedia.
\newblock In \emph{Proceedings of the international AAAI conference on web and social media}, volume~9, pages 454--463.

\bibitem[{Wiegreffe et~al.(2021)Wiegreffe, Marasovi{\'c}, and Smith}]{wiegreffe-etal-2021-measuring}
Sarah Wiegreffe, Ana Marasovi{\'c}, and Noah~A. Smith. 2021.
\newblock \href {https://doi.org/10.18653/v1/2021.emnlp-main.804} {{M}easuring association between labels and free-text rationales}.
\newblock In \emph{Proceedings of the 2021 Conference on Empirical Methods in Natural Language Processing}, pages 10266--10284, Online and Punta Cana, Dominican Republic. Association for Computational Linguistics.

\bibitem[{Xue et~al.(2020)Xue, Constant, Roberts, Kale, Al-Rfou, Siddhant, Barua, and Raffel}]{xue2020mt5}
Linting Xue, Noah Constant, Adam Roberts, Mihir Kale, Rami Al-Rfou, Aditya Siddhant, Aditya Barua, and Colin Raffel. 2020.
\newblock mt5: A massively multilingual pre-trained text-to-text transformer.
\newblock \emph{arXiv preprint arXiv:2010.11934}.

\end{thebibliography}

\appendix
\section{Fact Decomposition Prompt}
\label{sec:appendix:fact_decomposition}
We provide the fact-decomposition prompts in \texttt{fact\_decomp\_prompts.txt} at \url{https://github.com/smfsamir/infogap}. For \ru{}, we found that an example was required in order for the GPT-4 response to be consistently structured in the form of a \texttt{python} list of strings, while the other languages (\en{}, \fr{}) were able to successfully follow this instruction without an example.

\section{Fact Equivalence Prompt}
\label{sec:appendix:entailment_prompt}

We provide the prompts for \xfactmatch{} in \texttt{fact\_match\_prompts.txt} at \url{https://github.com/smfsamir/infogap}. The first row contains the prompt for \entoru{} and \entofr{}; the second for \frtoen{}; and the third for \rutoen{}. 
In each prompt, the \verb|src_facts| variable is equivalent to $e_{i-2}, e_{i-1}, e_{i}$ from \Sref{sec:x-fact-eq}, while \verb|tgt_facts| contains $f_{j-2},f_{j-1},f_{j}$, for $j\in [k]$. That is, we use these prompts to determine whether $e_i$ is contained in the other language (e.g., \fr{} for the \entofr{} direction).

\section{Seed biographies}
\label{appendix:biographies}
\begin{table}[]
    \centering
    \begin{tabular}{l}
    \entofr{}; \frtoen{} \\
    \toprule
        Gabriel Attal \\ 
        Ellen DeGeneres \\ 
        Tim Cook \\ 
        Kim Petras \\ 
        Alan Turing \\ 
        Caroline Mécary \\ 
        Abdellah Taïa \\ 
        Sophie Labelle \\ 
        Frédéric Mitterrand \\ 
        Philippe Besson \\ 
    \bottomrule
    \end{tabular}
    \caption{Initial seed set of people for obtaining \infogap{} labels with GPT-4 for the \entofr{} and \frtoen{} directions; see \Sref{sec:method:reliability}. We used the \infogap{} labels on this seed set to finetune a \texttt{flan-t5-large} model; see \Sref{sec:lgbt-implementation}.}
    \label{tab:en-fr-seed-set}
\end{table}

\begin{table}[]
    \centering
    \begin{tabular}{l}
    \entoru{}; \rutoen{} \\
    \toprule
        Pyotr Ilyich Tchaikovsky \\
        Tim Cook \\
        Dmitry Kuzmin \\
        Masha Gessen \\
        Nikolay Alexeyev \\
        James Baldwin \\
        Ali Feruz \\
        Elena Kostyuchenko \\
        Mikhail Zygar \\
        Pyotr Verzilov \\
        Sergey Sosedov \\
        Yekaterina Samutsevich \\
    \bottomrule
    \end{tabular}
    \caption{Initial seed set of people for obtaining \infogap{} labels with GPT-4 for the \rutoen{} and \entoru{} directions;  see \Sref{sec:method:reliability}. We used the \infogap{} labels on this seed set to finetune a \texttt{mt5-large} model; see \Sref{sec:lgbt-implementation}.}
    \label{tab:en-ru-seed-set}
\end{table}

In  \Table{tab:en-fr-seed-set} and \Table{tab:en-ru-seed-set}, we list the seed set of biographies, that were used for obtaining \infogap{} labels. We performed our human annotation experiment \Sref{sec:method:reliability} for \infogap{} on these labels. We then used these labels to distill \texttt{flan-t5-large} and  \texttt{mt5-large} for our analyses in \Sref{sec:lgbt_bio}.

We also used these seed biographies for obtaining connotation labels from GPT-4 for our analysis in \Sref{sec:lgbt_bio}, which we then also used to distill into \texttt{flan-t5-large} and \texttt{mt5-large} for predicting connotation labels at a larger scale.

\section{InfoGap Distillation Hyper-Parameters}
\label{sec:appendix:hyperparams}

\begin{table*}[]
    \centering
    \begin{tabular}{ll|ll}
    Model & Task & hyperparameter & value\\
    \toprule
      \texttt{flan-t5} &  Fact decomp. &  \verb|auto_find_batch_size| & True \\
         &  & Learning rate & 5e-5\\
         &  & Num. epochs & $5$\\
    \midrule
    & \xfactmatch{} & \verb|auto_find_batch_size| & True \\
    &  & Learning rate & 5e-5\\
    &  & Num epochs & 5\\
    \midrule
    & Conn. prediction & \verb|auto_find_batch_size| & True\\ 
    & & Learning rate & 5e-5\\
    & & Num. epochs & 5\\
    \bottomrule
    mT5  & Fact decomp. & Batch size & 2\\ 
    & & Learning rate & 9.5e-4\\
    & & Weight decay & $0.0$\\
    & & Gradient accumulation steps & 4\\
    & & Num. epochs & 5\\
    \midrule
    & \xfactmatch{} & Batch size & 2\\ 
    & & Learning rate & 8.5e-5\\
    & & Weight decay & $0.4$\\
    & & Gradient accumulation steps & 4\\
    & & Num. epochs & 5\\
    \midrule
    & Conn. prediction & \verb|auto_find_batch_size| & True\\ 
    & & Learning rate & 5e-5\\
    & & Num. epochs & 5\\
    \bottomrule
    \end{tabular}
    \caption{Parameters provided to the HugggingFace trainer for the \texttt{flan-t5-large} and \texttt{mt5-large} models.}
    \label{tab:hyperparams}
\end{table*}
We report fine-tuning hyperparameters for the HuggingFace Trainer in \Table{tab:hyperparams}. Unspecified values use the default setting of the Trainer (\texttt{python} version: 4.34.1). For all tasks, we used a train/test split of $0.9/0.1$. As for evaluation metrics, we used Rouge-1 for fact decomposition (\Sref{sec:x-fact-retrieve}),  and Micro-F1 for  \xfactmatch{} (\Sref{sec:x-fact-eq}) and connotation prediction (\Sref{sec:lgbt-rq2}). For the \entofr{} and \entofr{} directions, we apply the \texttt{flan-t5} models. We obtained strong validation set performance ($0.85$ Rouge-1; $0.85$ and $0.88$ F1s for the connotation prediction and \xfactmatch{} tasks, respectively) using the same hyperparameter settings across all three tasks.

We found that \texttt{flan-t5-large} did not generalize well to \ru{}, obtaining poor performance in fact decomposition and often predicting nonsensical Russian strings. We thus resorted to \texttt{mt5-large} instead \citep{xue2020mt5}, since Russian is one of the largest languages in terms of its pre-training data sizes. After a hyperparameter sweep over learning rates, gradient accumulation sizes, and weight decay values, we found much better performance with mT5, obtaining validation set performances of $0.89$, $0.79$, and $0.86$ for fact decomposition, connotation prediction, and \xfactmatch{} tasks, respectively.

All finetuning was completed on a single NVIDIA L40 GPU. 

\section{Connotation modeling} \label{appendix:connotation}


\paragraph{Forced alignment procedure.} As mentioned in \Sref{sec:lgbt-rq2}, we applied forced alignment to assign decomposed facts back into their original full sentences. Forced alignment is a constrained version of \href{https://en.wikipedia.org/wiki/Dynamic_time_warping}{Dynamic Time Warping}, where the alignment is monotonic. Forced alignment requires a distance function, we used hubness-corrected distance (\Sectref{sec:x-fact-retrieve}).

\paragraph{Connotation prompts.} We provide the prompts used for obtaining connotation labels in \texttt{connotation\_prompts.txt} at \url{https://github.com/smfsamir/infogap}. The \verb|content| variable contains up to 3 sentences, $s_{i-2}, s_{i-1},s_{i}$. While we're interested in the connotation towards \verb|person_name| conveyed in the last sentence $s_i$, we provide the prior two sentences for more context. 
We prompted for both connotation labels and rationales for the labels, after finding that prompting for a rationale prevented the models from vastly overextending the \texttt{neutral} label. This aligns with prior research on text classification, where generating rationales improved accuracy  \citep{wiegreffe-etal-2021-measuring}.   

\subsection{Validation of connotation label predictions}\label{appendix:connotation-validation}
\begin{table}[]
    \centering
    \begin{tabular}{cc}
         \textbf{Language} & \textbf{Macro-averaged F1}  \\
         \toprule
         \en{} & $0.77$\\
         \fr{} & $0.77$ \\
         \ru{} & $0.86$\\
         \bottomrule
    \end{tabular}
    \caption{Macro-averaged F1 scores for predicting the connotation towards the subject of a biography from a snippet of text in the biography. }
    \label{tab:connotation-validation}
\end{table}
To validate the connotation labels predicted in \Sref{sec:lgbt-rq2}, we sampled $10$ positive, $10$ negative, and $10$ neutral connotation label predictions from \Appendix{appendix:biographies} for each of the 3 languages, thus obtaining $90$ datapoints in total. One co-author then annotated each datapoint manually, and compared the annotations against the labels predicted by GPT-4 from the connotation prompt in \Appendix{appendix:connotation}. 

We provide the results of this classification in \Table{tab:connotation-validation}. We find that the connotation predictions are generally reliable, with all errors stemming from confusion between \texttt{neutral} and \texttt{positive}, or \texttt{neutral} and \texttt{positive}, rather than the more severe error of confusing \texttt{positive} and \texttt{negative} labels. This aligns with observations in previous research on computational modeling of connotation \citep[][among others]{park2021multilingual,rashkin2015connotation,sap2017connotation}. 

\paragraph{Distillation.} Having validated the quality of the GPT-4 connotation predictions, we use the predicted labels to finetune more scalable, lightweight models for predicting the connotation labels. We provide hyperparameter details in \Appendix{sec:appendix:hyperparams}.


\section{Regression model fitting} \label{appendix:bayesian-regression}
We fit the regression model using the \texttt{brms} package \citep{burkner2017brms}, with $2500$ steps (500 warmup) of the NUTS sampler \citep{hoffman2014no}. We used a regularizing $\mathcal{N}(0,10)$ prior on all the coefficients for the factors. 

\section{Identifying biographies with a positive connotation imbalance across language versions} \label{appendix:imbalanced-bio-test}
\paragraph{Plan.} We consider the \entofr{} direction for an arbitrary bio, without loss of generality. We will use the amount of neutral facts shared by both articles to parameterize a \texttt{BetaBinomial} distribution. After fitting this distribution, we will simulate draws from it to predict how much \textit{positive} information should be shared by both articles. When the actual amount of shared positive connotation facts  is much lower than the amount predicted by the fitted \texttt{BetaBinomial} distribution, we can consider this an imbalanced biography for the \entofr{} direction. 

\paragraph{Implementation.} We first set the prior for the neutral fact distribution to uniform (prior to observing the actual neutral overlap ratio): $\operatorname{Beta}(1,1)$. We leverage the useful fact that the posterior distribution after observing $x$ neutral facts $e_{i(1)}, \dots, e_{i(x)}$ in both \en{} and \fr{} out of $n$ total facts in \en{} is $\operatorname{Beta}(1+x,1+n-x)$ \citep{mackay2003information}. We can then simulate draws from the \texttt{BetaBinomial} distribution, first drawing a sample from $\operatorname{Beta}(1+x,1+n-x)$, followed by predicting amount of \en{} facts that \textit{should} also be found in \fr{}. The number of trials is fixed to the total number of positive facts in the \en{} article. 

Thus, this binomial distribution tells us the number of positive facts we would expect to see in both articles, if positive facts were not omitted at a higher rate than neutral facts. We can then draw $S=1000$ samples, counting the number of times $K$ the \textit{expected} amount of shared positive connotation facts is higher than the \textit{actual} amount. When $K / S$ is close to $1.0$, there is a large amount of positive information being omitted the \fr{} article, compared to the \en{} one. We use $1-K/S$ as a $p$-value, with an $\alpha=0.05$.

We emphasize further that this method can be applied in either direction (e.g., $\entofr{}$, or $\frtoen{}$), as well as for finding negatively imbalanced biographies, where one language version includes negative content at a rate much higher than expected under the neutral rate. 

\section{Hoeffding's inequality} \label{appendix:hoeffding}
We provide the full statement of Hoeffding's inequality for easy reference \citep{mitzenmacher2017probability,harvey2022first}: 
\begin{theorem}[Hoeffding's inequality] 
Let $X_1, \dots, X_n$ be independent random variables such that $X_i$ always lies in the interval $[0,1]$. Define $X=\sum_{i=1}^{n}X_i$. Then 
$\text{Pr}\left[ \lvert X-E[X]\rvert  \geq t\right] \leq 2\exp(-t^2/2n)$.
\end{theorem}

\end{document}